\documentclass[letterpaper, 10 pt, conference]{ieeeconf}

\usepackage{ifpdf}
\usepackage{cite}
\usepackage{graphicx}
\usepackage{caption}
\usepackage{amsmath}
\usepackage[ruled,linesnumbered]{algorithm2e}
\usepackage{algpseudocode}
\usepackage{array}
\usepackage{pseudocode}
\usepackage{fancybox}
\usepackage{comment}
\usepackage{amssymb}
\usepackage{xcolor}
\usepackage{soul}
\usepackage{romannum}
\usepackage{todonotes}
\usepackage{subfig}
\usepackage{commath}
\usepackage{mathrsfs}
\usepackage{stackengine}
\usepackage{multirow}
\usepackage{stmaryrd}
\usepackage{optidef}

\newtheorem{assumption}{Assumption}

\newtheorem{remark}{Remark}
\newtheorem{problem}{Problem}

\IEEEoverridecommandlockouts                           

\title{\LARGE \bf Probabilistically Resilient Multi-Robot Informative Path Planning}

\author{Remy~Wehbe and Ryan~K.~Williams  
\thanks{
R.~Wehbe and R.~K.~Williams are with the Department of Electrical and Computer Engineering, 
Virginia Polytechnic Institute and State University, Blacksburg, VA USA,
\mbox{E-mail: \textit{\{rewehbe, rywilli1\}@vt.edu}.}
}
}

\begin{document}

\maketitle
\thispagestyle{empty}
\pagestyle{empty}

\begin{abstract}

In this paper, we solve a multi-robot informative path planning (MIPP) task under the influence of uncertain communication and adversarial attackers. The goal is to create a multi-robot system that can learn and unify its knowledge of an unknown environment despite the presence of corrupted robots sharing malicious information. We use a Gaussian Process (GP) to model our unknown environment and define informativeness using the metric of mutual information. The objectives of our MIPP task is to maximize the amount of information collected by the team while maximizing the probability of resilience to attack. Unfortunately, these objectives are at odds especially when exploring large environments which necessitates disconnections between robots. As a result, we impose a probabilistic communication constraint that allows robots to meet \emph{intermittently} and resiliently share information, and then act to maximize collected information during all other times. To solve our problem, we select meeting locations with the highest probability of resilience and use a sequential greedy algorithm to optimize paths for robots to explore. Finally, we show the validity of our results by comparing the learning ability of well-behaving robots applying resilient vs.\ non-resilient MIPP algorithms. 

\end{abstract}

\section{Introduction}

In recent years, advancements in communication, processing capabilities, and hardware platforms, among others, have set the stage for huge interest in studying and controlling collaborating robots. Multi-robot systems (MRSs) bring forth several advantages in the form of redundancy, efficiency, and heterogeneity, with recent examples of application to important real-world problems \cite{Patnayak2021-mz,Santilli2022-ui,Liu2021-ll,Rangwala2021-vb,Heintzman2021-zr,Williams2020-sv,Heintzman2021-mw,Liu2018-xm,Liu2021-gw}. Unfortunately, MRSs face a unique set of challenges. Their performance is strongly related to coordination, a challenging task when operating in uncertain, noisy environments \cite{communication_noise, MRS_challanges_coordinate}, requiring complex algorithms to efficiently solve state-of-the-art problems \cite{MRS_challanges_complex_algo}. Additionally, MRSs are susceptible to adversarial attacks that threaten nominal operation \cite{MRS_challanges_attacks,leblanc_resilient,pasq_main}. Motivated by these limitations, we tackle the classical Multi-Robot Informative Path Planning (MIPP) problem, under the combined influence of uncertain communication and the presence of corrupted robots.

A powerful tool often used to model unknown environments is a Gaussian Process (GP) \cite{GP_book}. GPs are a statistical method for performing regression that can make predictions with a built in notion of uncertainty. As a result, GPs have been used extensively to represent unknown environments for informative path planning \cite{gp_ipp, gp_ipp2, gp_ipp3, krause_ipp, sukahtme_ipp}, where the task is to optimize robot measurements (information) while satisfying a set of constraints. In \cite{krause_ipp, krause_ipp_underwater}, the authors use GPs, the metric of mutual information, and the recursive greedy algorithm \cite{recursive_greedy_algorithm} to plan informative paths for a team of robots performing environmental monitoring and underwater exploration. In \cite{sukahtme_ipp}, the authors use a branch and bound algorithm to maximize variance reduction when planning informative paths. Another core theme is that robot coordination requires a certain level of communication between robots. In \cite{larkin_ipp} the authors tackle the multi-robot interaction planning problem by imposing an intermittent communication constraint on the path planner. While \cite{multi_robot_periodic_conn} tackles the problem of multi-robot coordination with periodic connectivity.


Coordination also requires that robots reliably and resiliently exchange information. Unfortunately, MRSs are often subject to adversarial attackers that aim to disrupt nominal operation. As a result, \cite{Bruno_paper, pasq_main} developed graph theoretic conditions for the detection and identification of malicious behavior in an MRS. In \cite{leblanc_resilient}, the authors tackle the problem of resilient consensus in large-scale MRSs by introducing the Weighted-Mean-Subsequence-Reduced (W-MSR) algorithm that only relies on local information. \cite{prorock_reselience_2} builds on the former work to present the sliding W-MSR algorithm (or SW-MSR) which allows dynamic MRSs, with time varying communication graphs, to achieve resilient consensus. Secure and resilient design requires that the communication graph of the MRS follows certain topological conditions. However, under uncertainty in communication, one cannot guarantee the existence of these topological properties. In \cite{tro_1, ral_2,Wehbe2019-cv,Wehbe2018-nh,Wehbe2019-la} the authors drop the assumption of deterministic communication and develop a probabilistic model to represent the topological conditions required for the design of a secure and resilient MRS under adversarial attack.


Prior work adopted MRSs with deterministic communication and individually tackled the problems of planning informative paths, planning under communication constraints, and planning with corrupted robots in the MRS. In this work, we aim to plan informative paths for an MRS to explore an unknown environment under the \emph{combined} influence of uncertainty in communication and the presence of adversarial attackers, which falls under open problems 2-4 presented in the very recent resilience taxonomy proposed in \cite{Resilienct_survey}. We model our unknown environment using Gaussian Processes, adopt a probabilistic communication model that can account for uncertainty, and use a resilient consensus algorithm to unify the robots' knowledge of the environment. The result is a multi-objective optimization problem where we aim to plan paths that maximize the amount of information collected while maximizing the probability that the robots are resilient against adversarial attackers. In large-scale environments learning about a process will necessitate disconnection and thus it becomes inconsequential to think about network resilience. At the same time, we eventually want all robots to have a unified understanding of the environment. Thus, we enforce an intermittent communication constraint where robots meet and resiliently share information with the objective of maximizing the amount of information collected during all other times. We solve our Resilient-MIPP problem by selecting meeting locations of highest probabilistic resilience and then optimizing paths that maximize the amount of collected information using a sequential greedy algorithm. We show the validity of our results by comparing the learning ability of well-behaving robots using resilient and non-resilient MIPP algorithms.

\section{Preliminaries}
\subsection{Graph Theory and MRS Modeling}

In this work, we will leverage two types of graphs. A deterministic graph is one where the nodes and edges are not subject to randomness, while a probabilistic graph is one where edges are associated with a probability distribution that represents the likelihood that a given edge exists in the graph. Probabilistic graphs are a useful mathematical tool to represent uncertainty, in our case, they are used to model uncertainty in communication between robots. A directed deterministic graph is denoted by $\mathcal{G(V,E)}$ where the set of vertices is given by $\mathcal{V}=\{v_1, v_2, ..., v_n\}$ and the set of edges is given by $\mathcal{E}=\{e_{ij}\; | \; v_i, v_j \in \mathcal{V} \}$. We use the notation $e_{ij}$ to denote a directed edge between nodes $i$ and $j$, while we use the notation $e_i$ to denote the $i^{th}$ edge in some set. We also denote by $\mathcal{N}_i^{in}=\{j\in\mathcal{V} \;|\; e_{ji} \in \mathcal{E}  \}$ the set of in-neighbors of node $i$. A directed probabilistic graph is denoted by $\mathscr{G(V,E)}$, where the vertex set is \emph{deterministic} $\mathscr{V}=\mathcal{V}$, and the edge set $\mathscr{E}$ is \emph{probabilistic}. Each edge $e_{ij} \in \mathscr{E}$ is associated with a probability of existence $p_{ij}$ that can model uncertainty, interference, etc. We do not make assumptions regarding the distribution of edge probabilities $p_{ij}$ except that we require edges to follow the independence equation given below.

 \begin{assumption}
 [Independent Edges]\label{assp:independent_edges}
 All pairs of edges $e_{1},e_{2} \in \mathscr{E}$ are independent such that $P(e_{1}\cap e_{2})=P(e_{1})\times P(e_{2})=p_{1}\times p_{2}$. 
 \end{assumption}

\begin{figure} [h]
\centering
\includegraphics[scale=0.18]{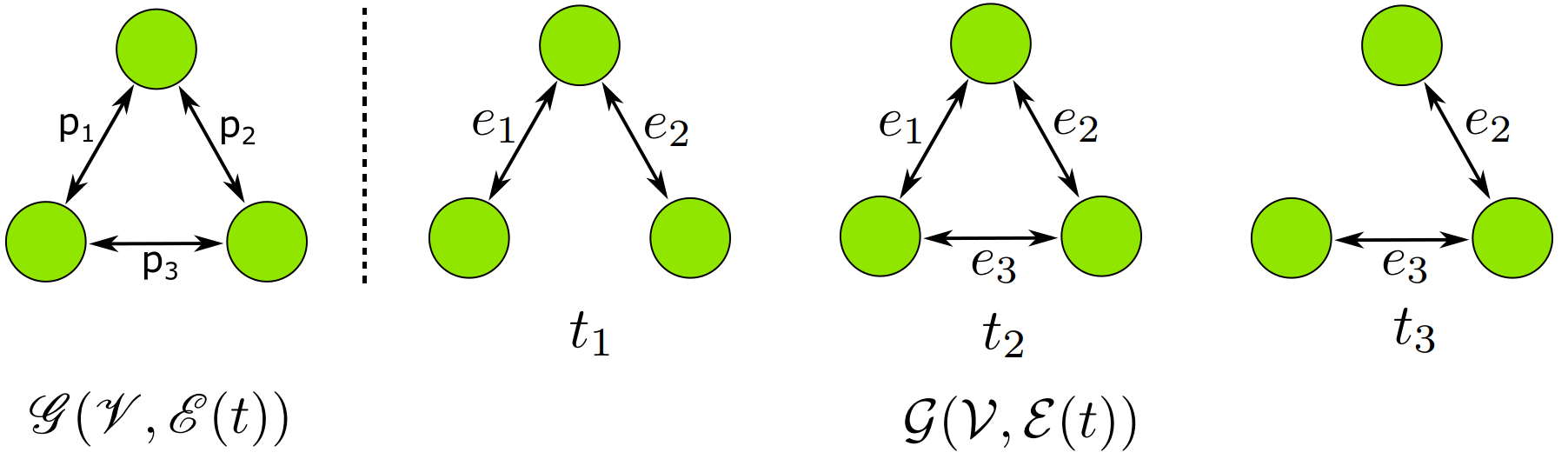}
\caption{Deterministic realizations of a probabilistic graph}
\label{fig:prob_to_det}
\end{figure}

In the above, we use the symbols $\cap$ and $\cup$ to represent AND/conjunction and OR/disjunction operations between events, respectively. To differentiate, we use the symbols $\wedge$ and $\vee$ to denote set intersection and union respectively. Our multi-robot system consists of a group of $n$ robots with indices in the set $\mathcal{I}_r=\{1,2,...,n\}$. Each robot is able to collect measurements $m_i^k$ of some environmental process of interest, i.e., the $i^{th}$ measurement collected by robot $k$. To model uncertainty in communication between robots we use a directed probabilistic graph $\mathscr{G(V,E}(t))$, where the time-dependence represents the dynamic behavior of an MRS. Notice however, that real MRSs evolve according to deterministic graphs. As a result, we adopt the deterministic realization model where the communication graph of the MRS is a series of deterministic graphs that evolve according to the probabilistic graph. In other words, the MRS evolves according to a time-varying deterministic graph $\mathcal{G(V,E}(t))$, that itself evolves according to an underlying $\mathscr{G(V,E}(t))$. Figure \ref{fig:prob_to_det} illustrates this concept. 
 
  We assume robots are operating in a discretized 2D space which is modeled using a directed graph $\mathcal{G}_d(\mathcal{V}_d,\mathcal{E}_d)$. Each vertex $v_i \in \mathcal{V}_d$ is a discrete location of interest and each edge $e_{ij} \in \mathcal{E}_d$ connects location $i$ to location $j$, where $\mathcal{C}(e_{ij})$ is the cost of moving from $i$ to $j$. Denote by $\ell_{ij}^k = \{e_1, e_2, ..., e_f\}$ a path from node $i$ to node $j$ associated with robot $k$ and by $\mathcal{C}(\ell_{ij}^k) = \overset{f}{\underset{i=1}{\sum}}\mathcal{C}(e_{i})$ the total cost of the path. In this work, we assume that robots have start and goal locations that are known a priori, as would be typical for example in a human-robot search and rescue application \cite{Heintzman2021-mw}.
	
\subsection{Gaussian Processes}

A Gaussian Process \cite{GP_book} is a powerful tool used for regression and classification that can make predictions with a built in notion of uncertainty. To model an environment using a GP, every location of interest $v_i \in \mathcal{V}_d$ is represented using a random variable $u_i$ where $\mathcal{L} =\{u_1, u_2, ...\}$ denotes the set of all locations of interest. Similarly, denote by $\mathcal{A} \subset \mathcal{L}$ the set of sensed locations in the environment and denote by $z_{\mathcal{A}}$ the vector of sensed values. A GP is fully described using a mean function $\mathcal{M}(u)$ and a covariance function $\mathcal{K}(u,v)$. Given a set of random variables $\mathcal{B}$, denote by $\mu_{\mathcal{B}}$ and $\Sigma_{\mathcal{B} \mathcal{B}}$ the mean vector and covariance matrix of the set $\mathcal{B}$. Then the set of equations that allow us to make predictions for any $y \in \mathcal{L}$ conditioned on the set $\mathcal{A}$ is given by \cite{krause_optimial_snsr_long}:
\begin{equation} \label{eq:posterior_mean}
f_{y|\mathcal{A}} = \mu_y +  \Sigma_{y\mathcal{A}} \Sigma^{-1}_{\mathcal{A}\mathcal{A}} (z_{\mathcal{A}} - \mu_{\mathcal{A}})
\end{equation}
\begin{equation} \label{eq:posterior_variance}
    \sigma^2_{y|\mathcal{A}} = \mathcal{K}(y,y) - \Sigma_{y\mathcal{A}}\Sigma^{-1}_{\mathcal{A}\mathcal{A}}\Sigma_{\mathcal{A}y} 
\end{equation}
where \eqref{eq:posterior_mean} and \eqref{eq:posterior_variance} are the posterior mean and posterior variance of the predicted locations. In this work, we adopt the common assumption of a zero mean function \cite{GP_book} and use the squared exponential as our covariance function,
\begin{equation}
    \mathcal{K}(u,v) = s_k^2\exp\bigg(-\frac{1}{2l_k^2}(u-v)^2\bigg)
\end{equation}
%
where $s_k$ and $l_k$ are the signal variance and length scale hyperparameters that define our covariance function. In informative path planning, the notion of informativeness can vary depending on the adopted metric. In this work, we select the typical metric of mutual information ($\mathcal{MI}$) \cite{krause_optimial_snsr_long} which seeks sensing locations that reduce the entropy of the space,
\begin{equation} \label{eq:mutual_info}
    \mathcal{MI(A)} = \mathcal{EN(L-A)} - \mathcal{EN((L-A)|A})
\end{equation}
where $\mathcal{EN(L-A)}$ is the marginal entropy and $\mathcal{EN((L-A)|A)}$ is the conditional entropy as defined in \cite{krause_optimial_snsr_long}.  Note that we have chosen well-known methods for process estimation (GP) and information representation ($\mathcal{MI}$), as our goal is to focus on the unique \emph{context} presented by uncertain communication combined with malicious attacks. 

\subsection{Resilient Consensus and Probabilistic Resilience}

Each robot $i\in \mathcal{I}_r$ in the MRS is associated with a state $x_i(t)$. This state can represent a measurement, velocity, direction, etc. Robots can achieve asymptotic consensus by updating their states according to local information received from their neighborhood. Specifically, robots can apply the following linear consensus protocol,
\begin{equation}\label{equ:linear_consen}
x_i(t+1)=w_{ii}(t)x_i(t)+ \underset{j \in \mathcal{N}^{in}_{i}}{\sum}w_{ij}(t)x_j(t)
\end{equation}
where the weights satisfy $w_{ij}>0$ and the sum of the weights adds up to one, i.e., $\sum_{ j \in i \cap \mathcal{N}^{in}_{i}} w_{ij}=1$. Essentially, each robot updates its state as a linear function of its own state and its neighboring robots' states. Unfortunately, the consensus algorithm given by \eqref{equ:linear_consen} is not resilient against the presence of misbehaving robots in the MRS, where a robot is said to be \emph{misbehaving} if it does not follow the update rule given by \eqref{equ:linear_consen}, or if it does not send the same value of its state to all of its neighbors. Indeed, \cite{Consenus_fail_single_attack} shows that a single misbehaving robot can drive consensus to an arbitrary value by keeping its state constant. A consensus algorithm is said to be \emph{resilient} if it allows well-behaving robots to achieve consensus in the presence of misbehaving robots. We adopt the Weighted-Mean-Subsequence-Reduced (W-MSR) algorithm \cite{leblanc_resilient} which is designed to achieve resilient consensus in the presence of up to $F$ misbehaving agents: At each time step $t$, each robot receives the state of the robots in its neighborhood. The robot proceeds to form a sorted list of its neighbors' states in increasing order. Next, if the list contains less than $F$ values that are larger than the robot's own state, the robot removes all of these values. If the list contains $F$ or more values that are larger than the robot's own state, only the largest $F$ values are removed. The same deletion rules are applied for the states in the list that are less than the robot's own state. Finally, the robot applies \eqref{equ:linear_consen} with the remaining states in the list.
 
For the W-MSR algorithm to converge, the communication graph for the MRS should satisfy the topological condition of $(r,s)$-robustness. Under the $F$-total model, which assumes the existence of up to $F$ corrupted robots in the MRS, the communication graph should be at least $(F+1, F+1)$-robust for the W-MSR algorithm to converge \cite{leblanc_resilient}. In deterministic graphs, one can check for the existence of the topological property of $(r,s)$-robustness \cite{leblanc_algo_robust_check, milp_robust_check}. Recall however, that our communication graph is probabilistic and thus we can only calculate the probability that the communication graph will satisfy the topological conditions of $(r,s)$-robustness, given by our previous work \cite{ral_2} as:
\begin{equation}\label{equ:prob_robust}
    P_{r}=P(\overset{|\mathfrak{S}|}{\underset{i=1}{\bigcap}}\overset{|\mathscr{R}_i|}{\underset{j=1}{\bigcup}} \mathcal{\acute R}_j)
\end{equation}
where $\mathfrak{S}$, $\mathscr{R}_i$, and $\mathcal{\acute R}_j$ are sets that are derived from the MRS's probabilistic communication graph. More specifically, $\mathfrak{S}$ is the set of all node pairs that satisfy the conditions of $(r,s)$-robustness, $\mathscr{R}_i$ is the set of edge sets that satisfy the conditions of $(r,s)$-robustness, and  $\mathcal{\acute R}_j$ is the event that an edge set satisfies the conditions of $(r,s)$-robustness. This equation allows us to make decisions regarding robot locations and configurations that have the highest likelihood of resilient information sharing. The probability represented by \eqref{equ:prob_robust} can be numerically calculated using a graphical method know as Binary Decision Diagrams. The interested reader is referred to \cite{ral_2} for a complete overview.

\section{Probabilistically Resilient Planning}
Robots in a team rely on the information they receive from their neighbors. Thus, if shared information is adversarial, one can expect the MRS's objective to either fail or at least deteriorate as a result. Additionally, an MRS operating in a real-world environment is subject to noise, interference, and obstructions that can severely affect the quality of communication between robots. In this paper, we tackle the problem of resilient environment exploration in the presence of adversarial attackers and probabilistic communication constraints. We formally define our problem as follows:

\begin{problem}

[Resilient-MIPP]: Given an MRS modeled using a probabilistic graph $\mathscr{G(V,E)}$, compute a motion plan that maximizes the amount of information collected while allowing well-behaving robots to have a unified understanding of the environment under the influence of probabilistic communication and adversarial attackers.
\end{problem}

\subsection{Attack Model and Team Objective}

We adopt the $F$-total \cite{leblanc_resilient} model where the total number of compromised robots in the system is bounded by $F$. Attackers are assumed to be  manipulating the corrupted robots' sensor measurements which is equivalent to a false data injection attack \cite{bruno_false_data_2} such that $m_i^{k,a} = m_i^k + \epsilon_i$, where  $m_i^{k,a}$ is the attacked sensor measurement and $\epsilon_i$ is the attack input on the $i$th measurement. Importantly, despite the existence of low-level sensor attacks, we assume that high-level decision-making is not compromised, i.e., all robots will execute the motion planning algorithm that we propose.

Next, we provide a high-level overview of the cooperative objective that the MRS is aiming to achieve. All robots start with the same GP that represents the unknown environment which can be built using a limited pilot deployment, expert knowledge, or can be completely unknown. The robots then proceed to periodically collect measurements aimed to maximize knowledge of the environment, then share the gathered information to unify their understanding of the unknown environment. During this process, robots are subject to probabilistic communication and may be compromised by corrupted robots that are sharing adversarial information.

Generally speaking, to maximize information collected as a team, each robot will measure a unique part of the environment. Unfortunately, the measurements of the attacked robots will be corrupted and since each location of interest is most likely measured by a single robot, there is no straightforward way to differentiate between a good and corrupted measurement. As as result, if robots choose to directly share their measurements with their neighbors, well-behaving robots will be corrupted by adversarial robots and the overall objective will fail \cite{gp_learn_under_attack} since the W-MSR algorithm can only be applied to data that is common to all robots. So, how should the robots resiliently share measurement information when each robot measures a different part of the environment? Although individual robot measurements are different, all robots are trying to learn the same GP which is mainly defined through its covariance function. Then the robots can first collect measurements, use these measurements to update the covariance function of their GP, then apply the resilient W-MSR algorithm on the shared hyperparameters $s_k$ and $l_k$ which define the covariance function. Note that the hyperparameters of corrupted robots will still be corrupted as they are computed from corrupted measurements. However, the W-MSR algorithm is now able to filter out this corrupted data for well behaving robots, thus allowing these robots to unify their knowledge of the environment based on the information collected by their neighbors while still being resilient.

\subsection{Modeling Probabilistic Communication}\label{subsec:prob_comm_resilience}

\begin{figure}[t!]%
    \centering
    \subfloat[]{{\includegraphics[width=4cm, draft = false]{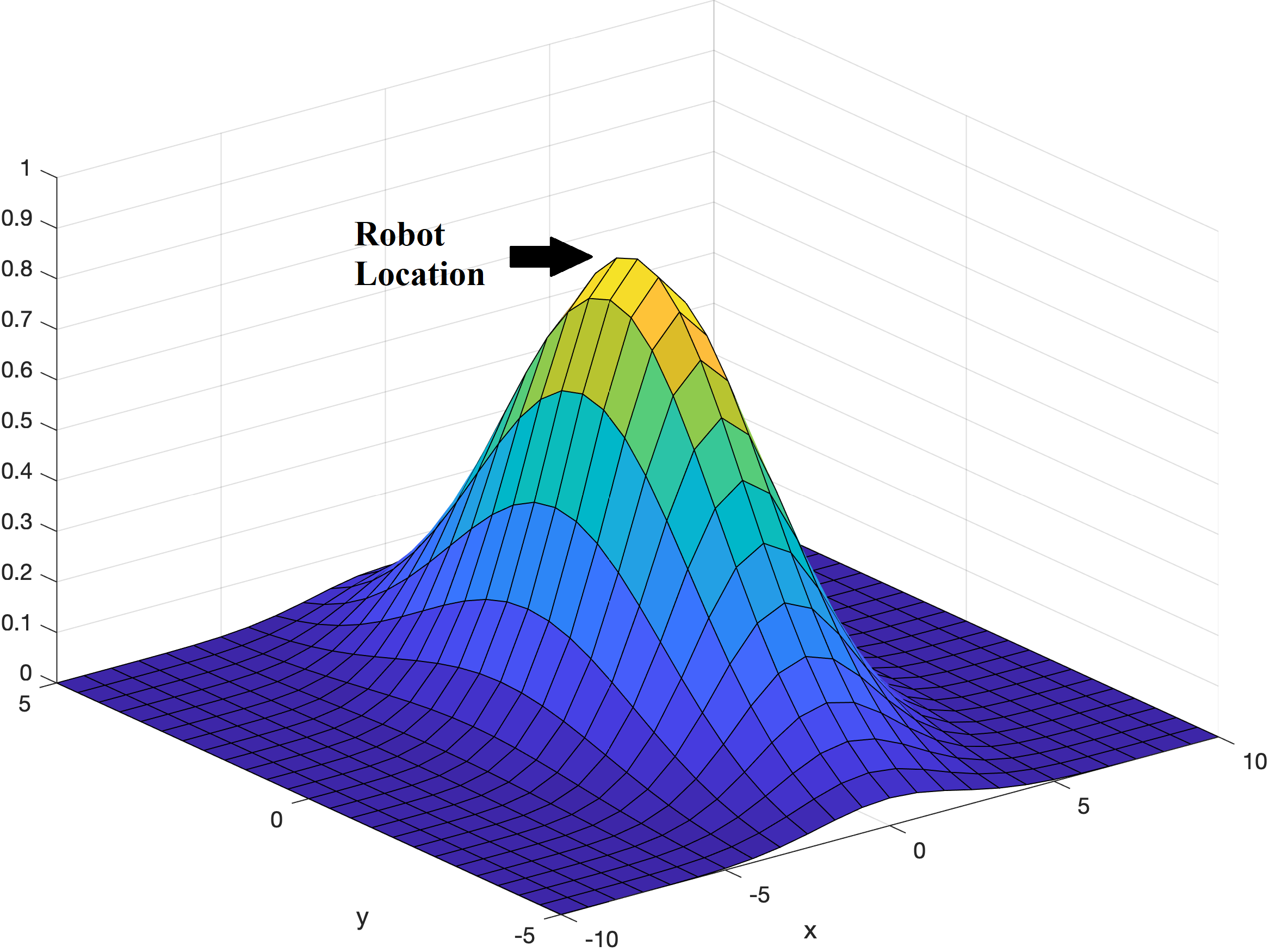} }\label{fig:robot_com_2}}%
    \smallskip
    \hfill
    \subfloat[]{{\includegraphics[width=4cm, draft =false]{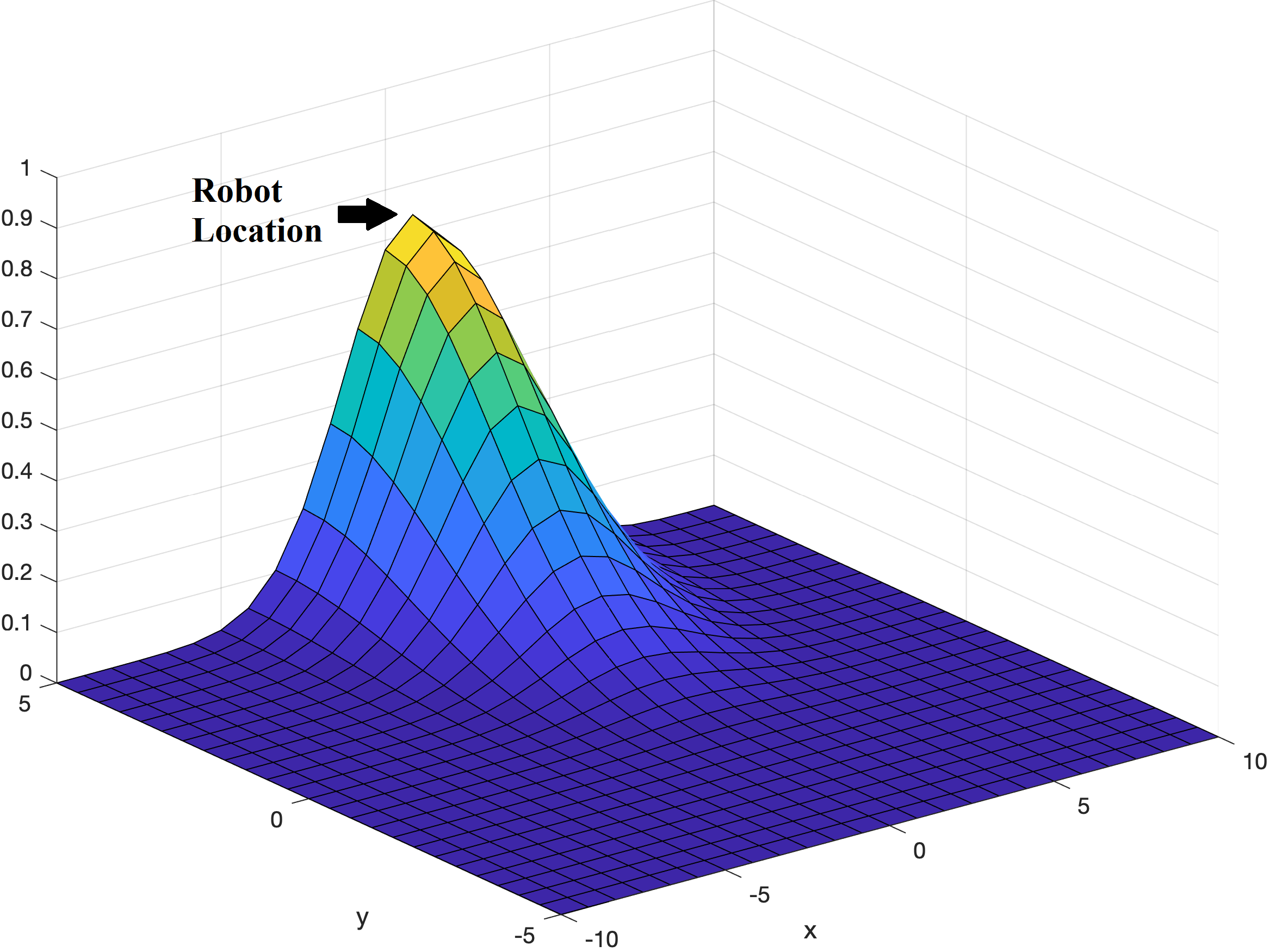} }\label{fig:robot_com_3}}%
        \caption{GPs modeling a robot's communication. (a) is for a robot located in the middle of the environment, (b) is for a robot located on the left side.}%
    \label{fig:gp_comms_ex}%
\end{figure}

Uncertainty in communication can result from many different influences such as obstructions, areas of high interference, and/or areas that may be adversarial. It was shown in \cite{krause_optimal_snsr_comm} that the quality of communication in an environment can be learned using a Gaussian Process. In this work, our main focus is to explore an unknown environment where probabilistic communication is a constraint which requires the communication GPs to be learned before the informative path planning task is initiated (this approach was also used in \cite{krause_optimal_snsr_comm}). As a result, we associate each location in the environment with a GP that models the probability of communication between that location and all other locations in the environment similar to what is shown in Figure \ref{fig:gp_comms_ex}. Additionally, we assume that these communication GPs are known a priori and will be used to calculate the probability of resilience of an MRS in the environment, i.e.,
\begin{equation}
    p_{ij} = \mu_j +  \Sigma_{j\mathcal{A}_i} \Sigma^{-1}_{\mathcal{A}_i\mathcal{A}_i} (z_{\mathcal{A}_i} - \mu_{\mathcal{A}_i})
\end{equation}
where $\mathcal{A}_i$ defines the communication GP at location $i$, and $p_{ij}$ is the probability that a robot located at node $i$ is able to communicate with a robot located at node $j$. Note that for an MRS to execute a complicated task such as MIPP, the robots need to exchange all required information. Since communication is probabilistic, we cannot guarantee that all the required information is directly shared from a single communication round. Rather, we relate the probability of communication to the expected number of re-transmissions, where high probabilistic resilience will result in fewer communication rounds while low probabilistic resilience will result in more communication rounds and thus less effective information sharing. We study this relationship in the simulations section.
	
\subsection{Optimization Problem for Resilient MIPP}
	
The first step is to formally define our multi-objective function. Since the unknown environment is modeled as a Gaussian Process where each location of interest $v_i \in \mathcal{V}_d$ is associated with a random variable $u_i \in \mathcal{L}$, a path $\ell_{ij}^k$ can be written in terms of the random variables associated with the nodes in the path such that $\mathcal{P}^k_{ij} = \{u_1, u_2, ..., u_f\}$. Then our optimization problem is defined as follows:
\begin{maxi}|s|
{\bar{\mathcal{P}}_{st}^k \subseteq \mathcal{L} \; \forall \; k}{\alpha_1\mathcal{MI}\bigg(\overset{n}{\underset{k=1}{\bigvee}}\mathcal{\bar{P}}_{st}^k\bigg) +\alpha_2 P(\overset{|\mathfrak{S}|}{\underset{i=1}{\bigcap}}\overset{|\mathscr{R}_i|}{\underset{j=1}{\bigcup}} \mathcal{\acute R}_j)}
{ \label{eq:mipp_optmz_cost}}{}
\addConstraint{\overset{n}{\underset{k=1}{\sum}}\mathcal{C}(\ell_{st}^k) \leq \gamma, \; \gamma \in \mathbb{Z}_{+}}
\end{maxi} 
where $\alpha_1$ and $\alpha_2$ are weighting parameters, and the subscript $st$ defines the source and terminal nodes of the path. Denote by $\mathcal{OPT}_1$ the first part of our optimization related to mutual Information, and denote by $\mathcal{OPT}_2$ the second part of our optimization related to the probability of resilience. It may not be immediately apparent that these optimizations depend on one another. However, recall that the paths selected by the robots, $\mathcal{P}_{st}^k$, induce a communication graph for the MRS. Based on this communication graph, the sets $\mathfrak{S}$, $\mathscr{R}_i$, and $\mathcal{R}_j$ are defined which in turn affects the probability of resilience. 

\begin{figure}[t!]%
    \centering
    \subfloat[]{{\includegraphics[width=4cm, draft = false]{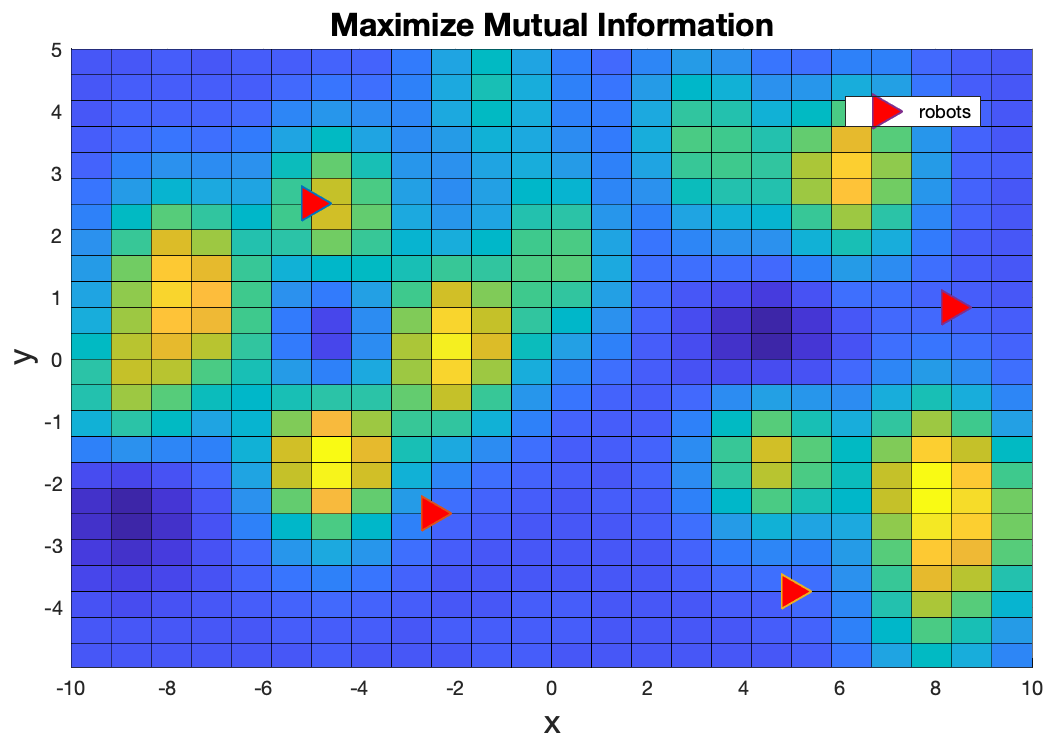} }\label{fig:max_mi}}%
    \smallskip
    \hfill
    \subfloat[]{{\includegraphics[width=3.8cm, draft = false]{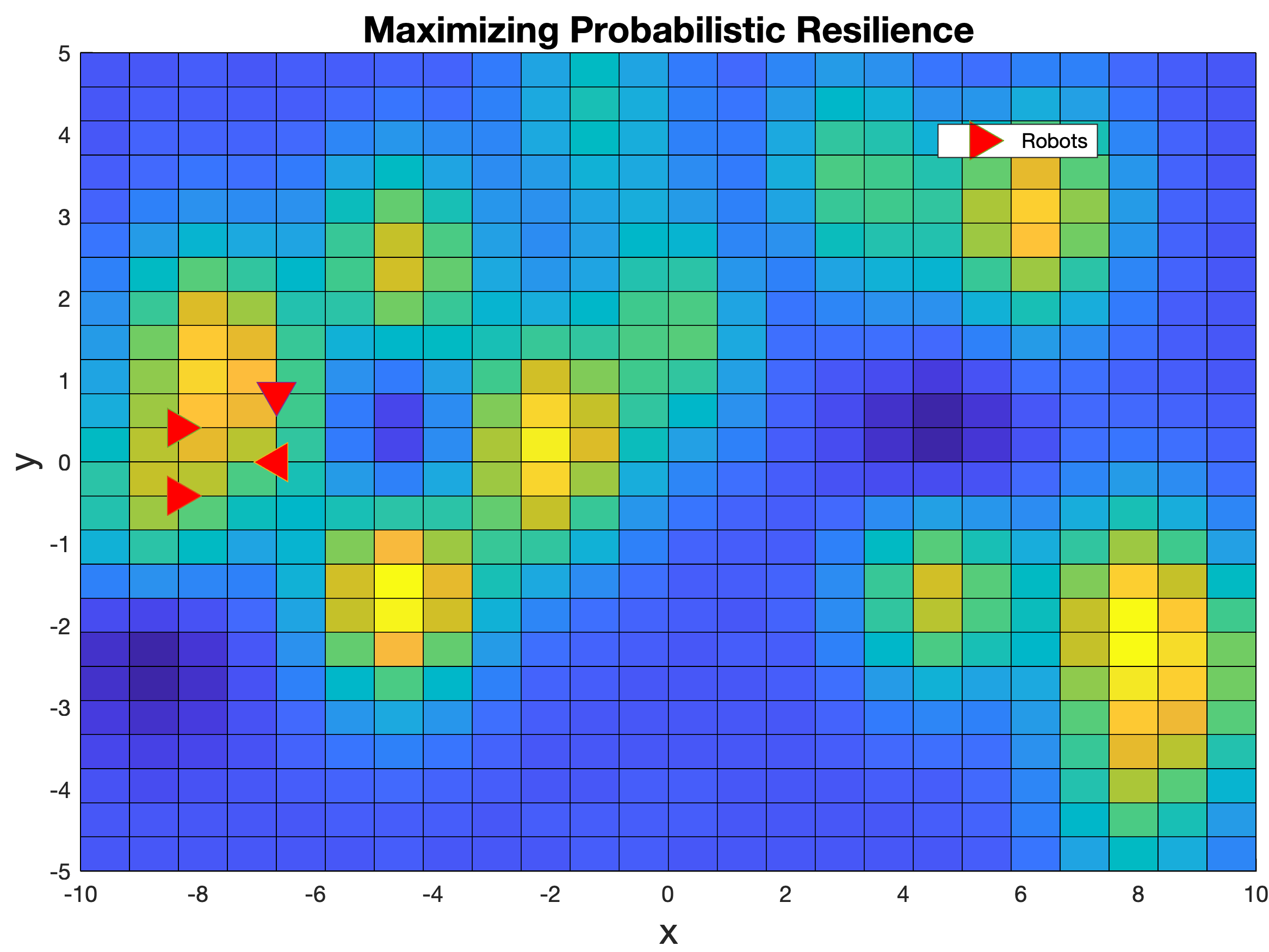} }\label{fig:max_res}}%
        \caption{Maximizing Mutual Information (a) vs. Probabilistic Resilience (b).}%
    \label{fig:max_mi_res}%
\end{figure}

Essentially, problem \eqref{eq:mipp_optmz_cost} aims to maximize the amount of information collected by the team of robots ($\mathcal{OPT}_1$) and at the same time maximize the probability that the MRS has a configuration that is resilient against adversarial attackers ($\mathcal{OPT}_2$), while restricting the total cost of exploration. Satisfying such a multi-objective function is hard because $\mathcal{OPT}_1$ and $\mathcal{OPT}_2$ are at odds. Maximizing $\mathcal{OPT}_1$ requires (generally) that the robots spread in the environment as much as possible, whereas maximizing $\mathcal{OPT}_2$ requires that the robots travel in areas of high probabilistic communication while staying in close proximity at all times as shown in Figure \ref{fig:max_mi_res}. In large-scale environments, learning about a process will necessitate disconnection and thus it becomes inconsequential to think about network resilience. At the same time, we eventually want all robots to have a unified understanding of the environment and thus we enforce information sharing intermittently where resilience of the network becomes a priority. As a result, we decouple the joint optimization problem above into subproblems of selecting a sequence of subareas (defined in the sequel) where each subarea maximizes probabilistic resilience so robots can meet and resiliently share information. That is:

%
\begin{maxi}|s|
{sb \in \mathcal{SB} }{P(\overset{|\mathfrak{S}|}{\underset{i=1}{\bigcap}}\overset{|\mathscr{R}_i|}{\underset{j=1}{\bigcup}} \mathcal{\acute R}_j)}
{ \label{eq:optz_res}}{}
\end{maxi} 	
Note that our optimization variable $sb$ does not directly show up in our optimization equation. Rather, similar to \eqref{eq:mipp_optmz_cost}, choosing the meeting location $sb$ induces a probabilistic communication graph which in turn determines the variables of our optimization equation. Then, during all other times, we select paths that maximize the amount of information collected by the team,
\begin{maxi}|s|
{\bar{\mathcal{P}}_{st}^k \subseteq \mathcal{L} \; \forall \; k}{\mathcal{MI}\bigg(\overset{n}{\underset{k=1}{\bigvee}}\mathcal{\bar{P}}_{st}^k\bigg)}
{ \label{eq:optz_ipp}}{}
\addConstraint{\overset{n}{\underset{i=1}{\sum}}\mathcal{C}(\ell_{st}^k) \leq \gamma, \; \gamma \in \mathbb{Z}_{+}}
\end{maxi} 

\subsection{Algorithm for Resilient MIPP}

\begin{figure*}[t!]%
    \centering
    \subfloat[Discretized areas]{{\includegraphics[width=5cm, draft = false]{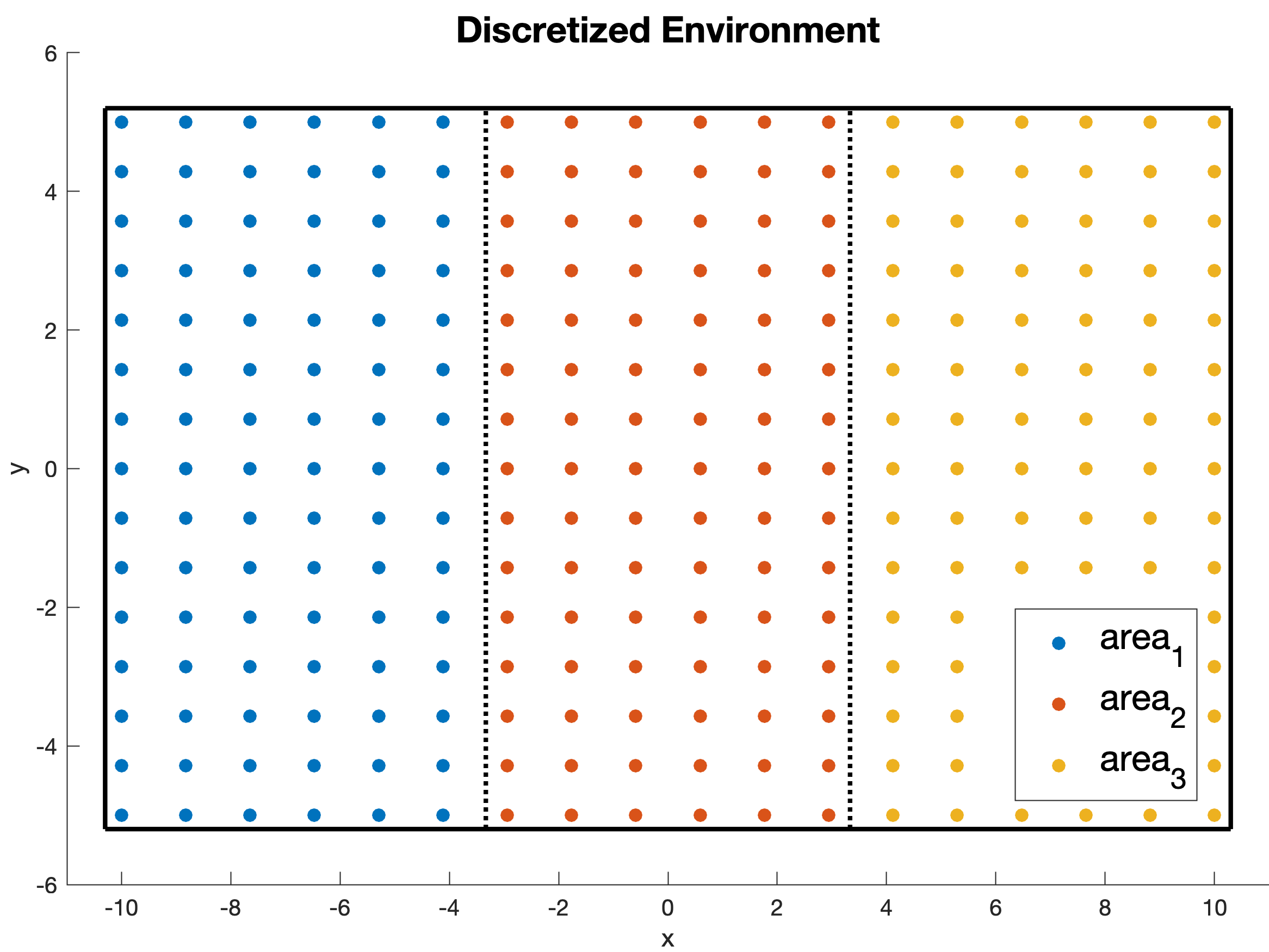} }\label{fig:discrete_area}}%
    \smallskip
    \hfill
    \subfloat[Subareas for robots to meet]{{\includegraphics[width=5cm, draft = false]{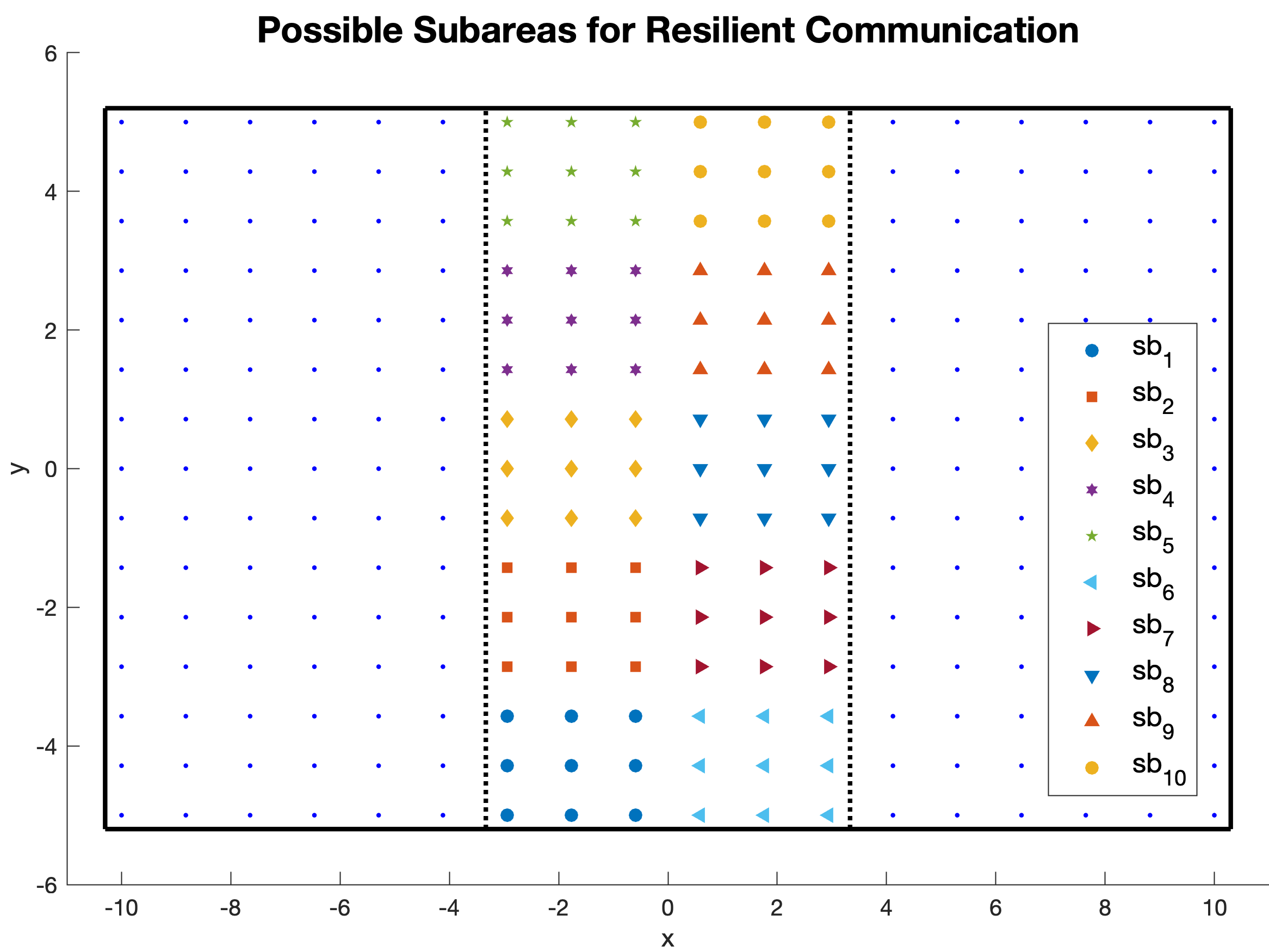} }\label{fig:discrete_subarea}}%
    \smallskip
    \hfill
    \subfloat[Initial map provided for the robots]{{\includegraphics[width=5cm, draft = false]{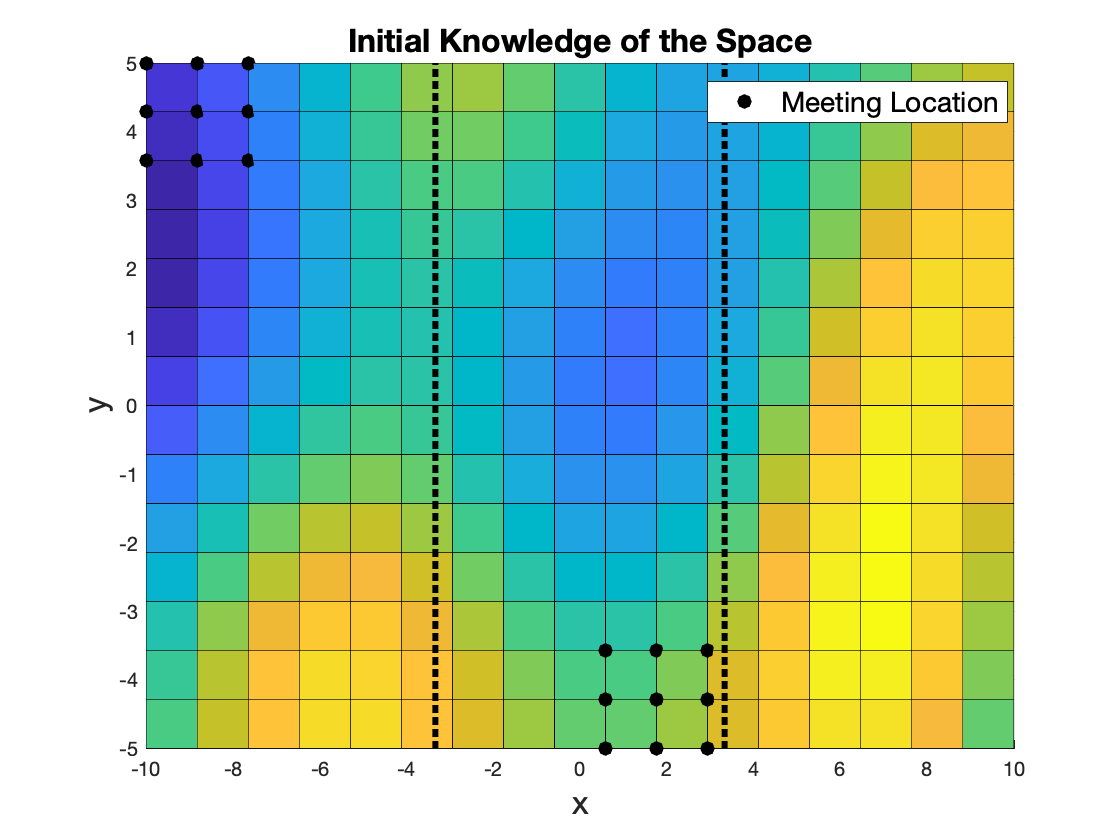} }\label{fig:discrete_init}}%
        \caption{Dividing the environment into areas and subareas for robots to explore and meet}%
    \label{fig:discrete_space}%
\end{figure*}

Optimization \eqref{eq:optz_res} is a combinatorial optimization problem which is often intractable to compute exactly. As a result, we will use a greedy approach to find the best subareas for robots to meet. To keep the problem structured, divide the environment into $\mathfrak{m}$ equally spaced areas denoted by $\mathcal{AR} = \{area_1, area_2, ..., area_{\mathfrak{m}}\}$. Figure \ref{fig:discrete_area} shows an environment divided into $\mathfrak{m} = 3$ areas \footnote{The method of dividing the environment into areas to explore will affect the paths chosen by the robots. In this work, we resort to the intuitive approach of dividing along the natural direction of motion since robots start from the left hand side and move to the right hand side of the environment. The division should be tailored according to the given task.}. For each $area_i \in \mathcal{AR}$, further divide the area into subareas $\mathcal{SB} = \{sb_1, sb_2, ..., sb_f\}$ as shown in Figure \ref{fig:discrete_subarea}. For each subarea, compute the probability of resilience using \eqref{equ:prob_robust} and the GP communication models (similar to the models shown in Figure \ref{fig:gp_comms_ex}). Then, the subarea with highest probabilistic resilience will be the meeting location of the robots. Note that for the last area, there is no need to compute a meeting location since the robots will meet at the end location of the environment. Figure \ref{fig:discrete_init} shows the initial environmental map provided to the robots at the beginning of the task for $\mathfrak{m} = 3$, along with the locations that robots will meet to exchange information. 


Now the problem has been reduced to computing an MIPP algorithm to explore each $area_i \in \mathcal{AR}$, subject to an implicit constraint induced by robots meeting at the optimizing subareas. Note that the previous computations are done offline and the resulting information is provided to the robots at the beginning of the task. Planning paths however is done online because as the robots gain new information about the environment, they can select better paths that maximize their knowledge of the unknown process. Several algorithms have been introduced to maximize the criterion of mutual information. Examples include the recursive greedy algorithm \cite{recursive_greedy_algorithm}, the efficient sequential allocation algorithm \cite{krause_ipp}, and a branch and bound approach \cite{sukahtme_ipp}. The choice of MIPP algorithm does not affect the core results of this paper as we are mainly concerned with learning an environmental process despite the presence of corrupt robots and uncertainty in communication. As a result, the user can select whichever MIPP algorithm best suits the application in terms of complexity vs. efficiency. In this work, we will introduce a simple greedy algorithm with a polynomial complexity that sacrifices some guarantees provided by the previous algorithms in favor of speed and ease of execution.
	
\begin{figure*}[t!]%
    \centering
    \subfloat[Path for robot 1]{{\includegraphics[width=4cm, draft = false]{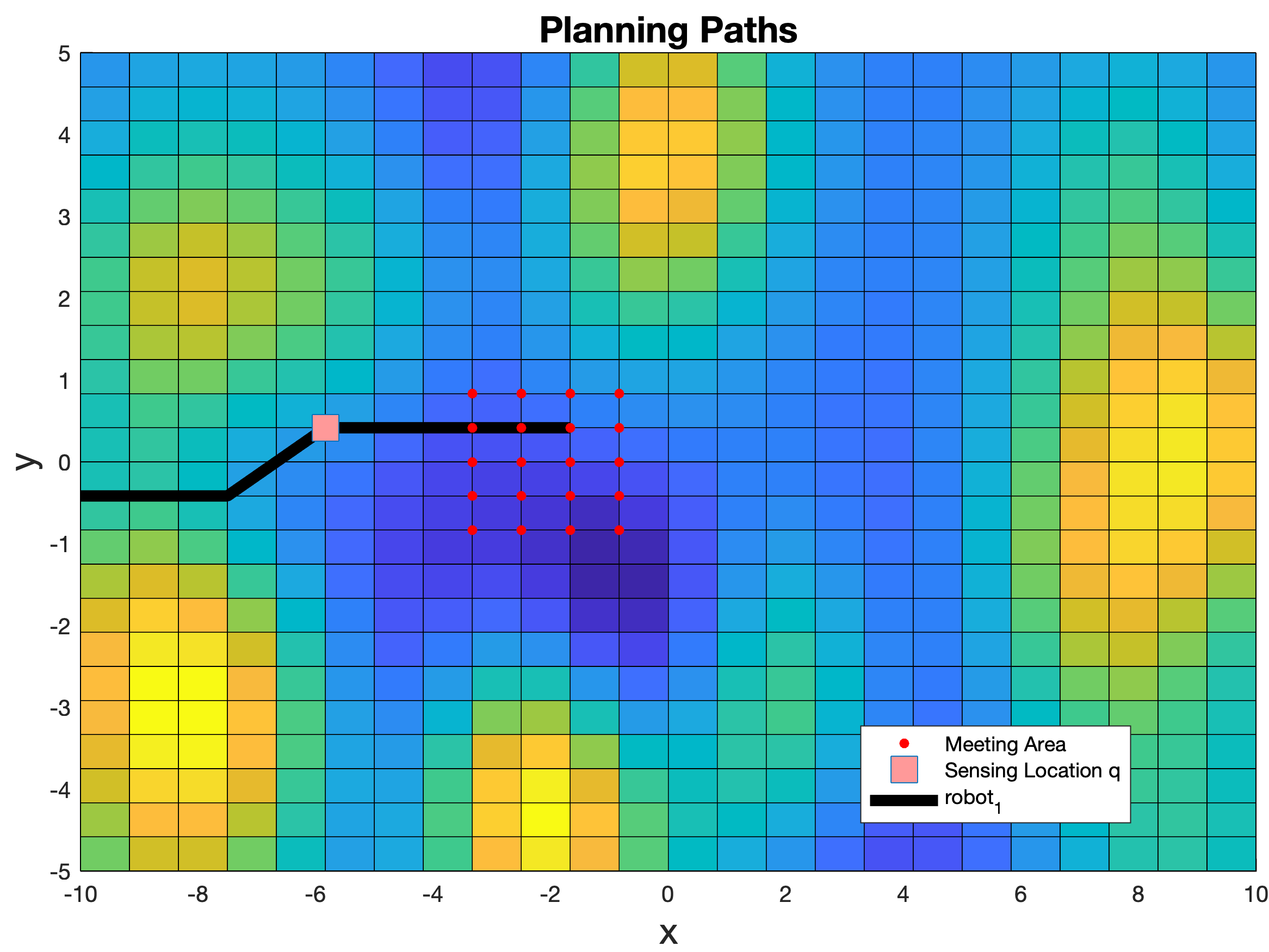} }\label{fig:path1}}%
    \smallskip
    \hfill
    \subfloat[Path for robot 2]{{\includegraphics[width=4cm, draft = false]{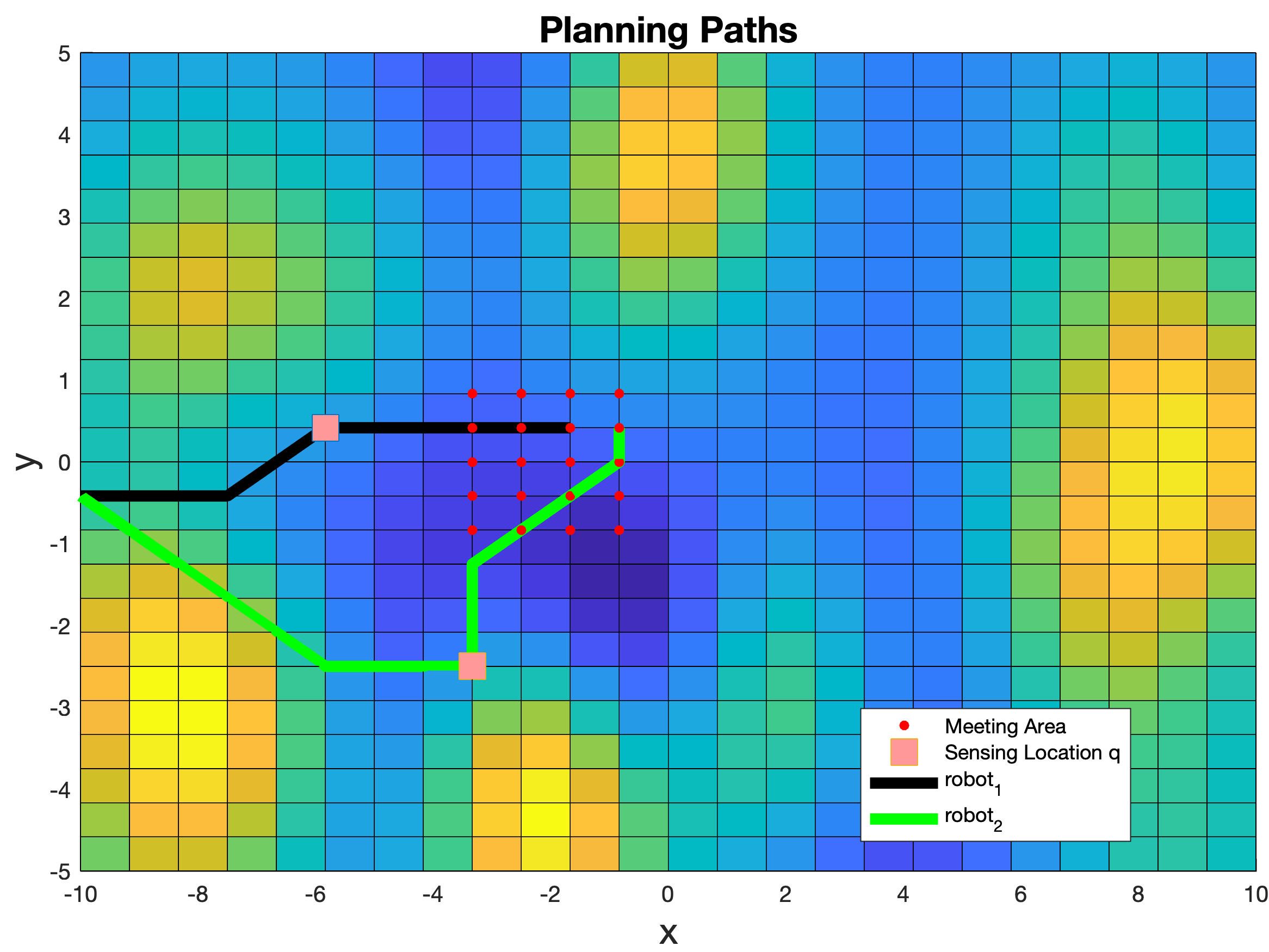} }\label{fig:path2}}%
    \smallskip
    \hfill
    \subfloat[Path for robot 3]{{\includegraphics[width=4cm, draft = false]{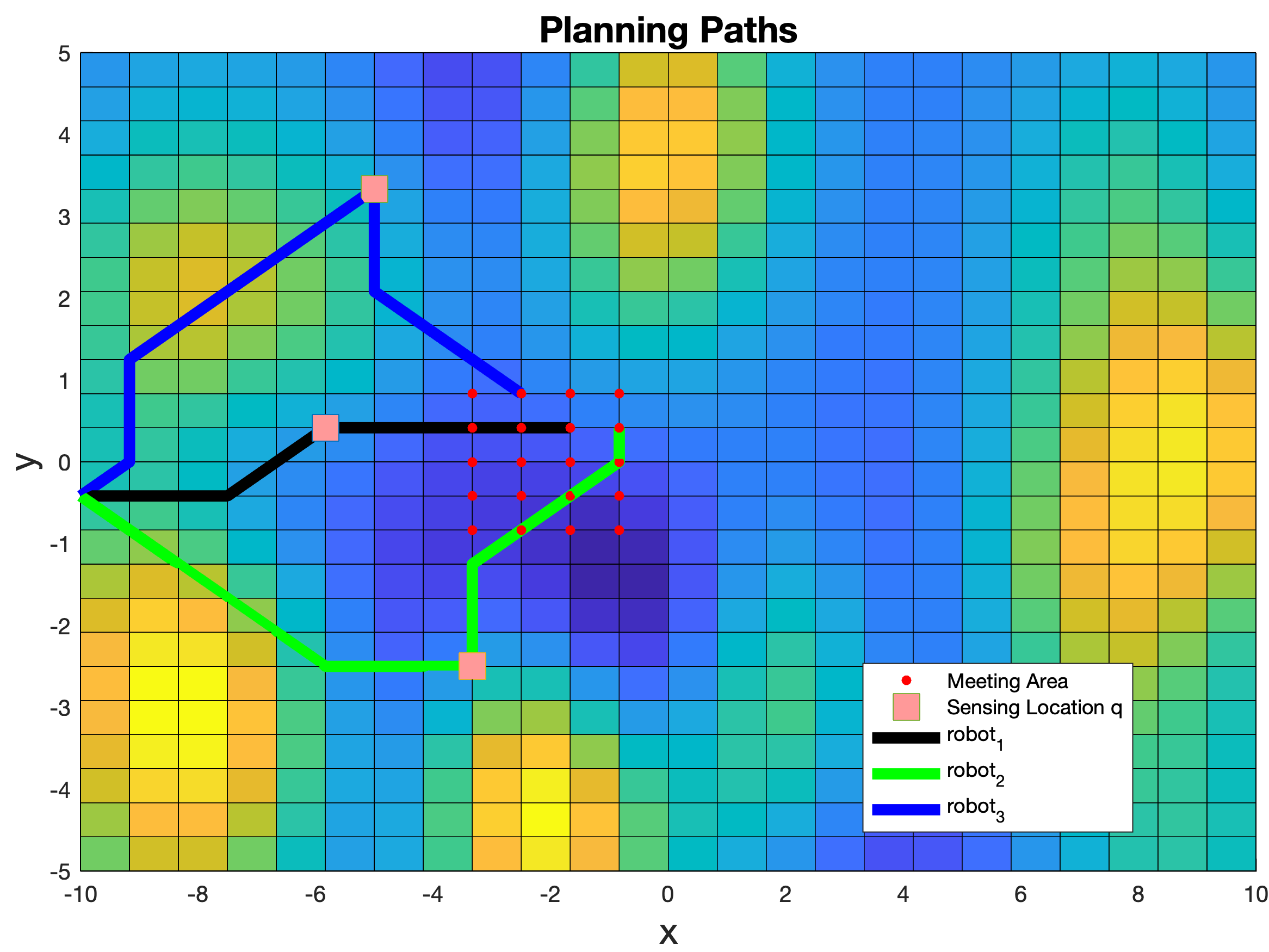} }\label{fig:path3}}%
    \smallskip
    \hfill
    \subfloat[Re-planning after meeting]{{\includegraphics[width=4cm, draft = false]{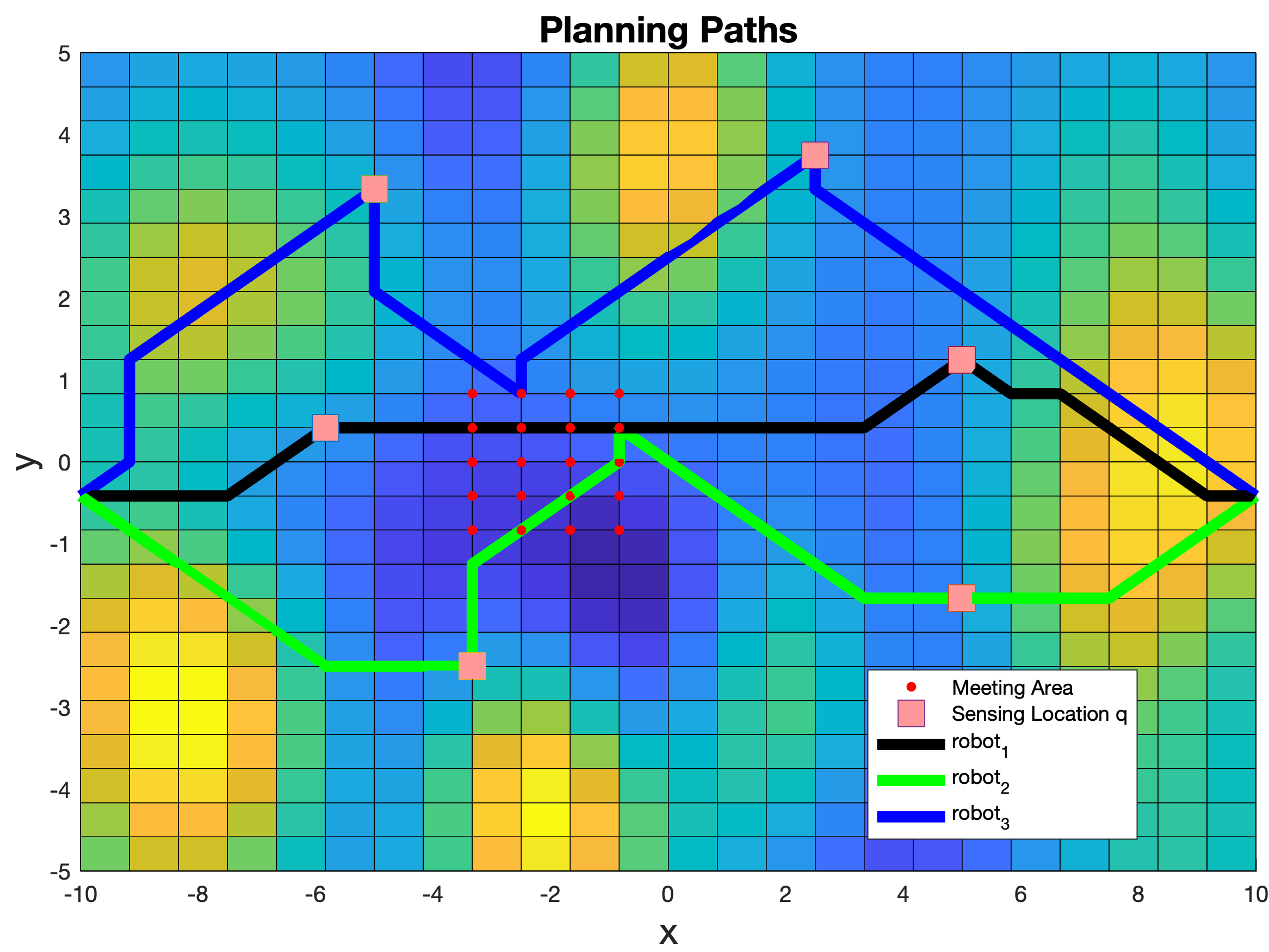} }\label{fig:path4}}%
    
        \caption{Planning informative paths for the MRS}%
    \label{fig:path_planning}%
\end{figure*}

The algorithm uses a sequential approach where initially, no robot has selected any path to explore. Then, one at a time, each robot plans a path $\mathcal{P}_{st}^k = \mathcal{P}_{sq}^k \vee \mathcal{P}_{qt}^k$ where $s$ is the current location, $q$ is an intermediate location, and $t$ is the end location. The starting and ending locations depend on the current location of the robot and the next meeting location, respectively. More importantly, we focus on selecting the intermediate location which will dictate the path taken by the robot and the quality of information collected. To this end, we select $q$ as an instance of the sensor placement problem \cite{krause_optimal_snsr_short} where the intermediate location of the robot is the location with highest mutual information in $area_i \in \mathcal{AR}$ that is going to be explored. To minimize travel cost, the $sq$ and $qt$ paths are computed using Dijkstra's shortest path algorithm. All points along the resulting path will be sensed by the robot as it moves in the environment. To avoid redundant exploration along the shortest path, we increase the weight of all edges that have been traversed using $\mathcal{C}(e_{ij}) = \alpha \mathcal{C}(e_{ij})$. The end result is an informative path for a single robot to follow. After the current robot plans its path, it shares its $\mathcal{P}_{st}^k$ path with all other robots in the MRS and the next robot plans its path accordingly (There is no need to secure the path sharing process since our attack model is restricted to sensor based attacks and thus the robots will share the actual path they will follow). This process is sequentially repeated until all the robots have a path planned. It is important to note that our algorithm is executed in real-time and the paths of each robot are re-planned at each meeting location. Notice that all robots start with the same GP representing the environment, however, as they collect information, each robot builds its own GP based on the measurements it has collected. Meeting and executing resilient consensus on the GP parameters allows all the well-behaving robots to unify their knowledge of the environmental GP without directly sharing measurements. Figure \ref{fig:path_planning} shows the sequential planning process for each robot when $\mathfrak{m} = 2$ and the re-planning process after meeting and exchanging information. Finally, the entire resilient MIPP algorithm is summarized in Algorithm \ref{alg:resilient_mipp}.

\begin{algorithm} 
\SetAlgoLined
\LinesNumbered
\small

 \textbf{Offline Computations:}\\
{ Divide the space into $\mathfrak{m}$ areas $\mathcal{AR} = \{area_1, ..., area_{\mathfrak{m}}\}$\;
  \ForAll{$area_i \in \mathcal{AR}-1$}{
   Divide $area_i$ into subareas $\mathcal{SB} = \{sb_1, ..., sb_f\}$\;
   \ForAll{$sb_j \in \mathcal{SB}$}{
   Robots get random locations in the meeting area\;
   Calculate the probability of resilience of the MRS\;}    
   Pick the highest probability of resilience meeting area\;}}

\textbf{Online Computations:}\\

\uIf{Location != Start Location}{
Each robot updates its GP based on the collected information\;
The robots execute the W-MSR Algorithm on GP covariance parameters\;
}

\For {k=1:n}{
$robot_k$ picks the sensing location of highest $\mathcal{MI}$\;
$robot_k$ plans a path $\mathcal{P}_{st}^k = \mathcal{P}^k_{sq} \vee \mathcal{P}^{k}_{qt}$\;
$robot_k$ shares planning information\;
} 
Robots execute the agreed upon plan\;
Repeat from step 12 until the end of the environment.

 \caption{Resilient MIPP}\label{alg:resilient_mipp}
\end{algorithm}

\section{Simulations}

In this section, we want to investigate the effects of corrupted robots on the learning capabilities of well-behaving robots when the MRS is performing an environmental MIPP task. Furthermore, we want to study how probabilistic resilience affects the number of communication rounds between the robots. As a result, we assume that our MRS is formed of $n$ robots out of which $n_a$ robots are being compromised. To evaluate the effectiveness of our proposed approach on the resilience of an MRS, we run resilient and non-resilient MIPP algorithms with a varying number of robots and attackers while recording the statistics that represent the well-behaving robots' ability to learn. To evaluate the effect of selecting high-resilience meeting locations, we study the average number of re-transmissions required for robots to fully share the required information when optimizing subareas of high probabilistic resilience versus random meeting locations. 

\begin{remark}
Previous work \cite{leblanc_resilient, prorock_reselience, prorock_reselience_2} have already established that the W-MSR algorithm will be effective at filtering out corrupted data when applied directly to the measurements exchanged. We note however, that in this work, we are applying the W-MSR algorithm in a new context by sharing information that is affected by the corrupted information (GP parameters), but not the information itself (sensing measurements). We show that our proposed approach of resiliently sharing the learned parameters combined with each robot's local measurements is able to resiliently unify the team's knowledge of the learned GP in the context of MIPP.
\end{remark}


\begin{remark}
The probability of resilience refers to the probability that in a single round of communication, the MRS is able to resiliently exchange the required information. However, if we restrict our MRS to a single round of communication, it will be very hard for the W-MSR algorithm to converge. We already know that the W-MSR algorithm can be applied using multiple rounds of communication \cite{prorock_reselience_2}. So as a result, we associate the cost of meeting at a bad location (low probability of resilience) with the expected number of re-transmissions needed to achieve resilient consensus. This is a necessary approach to allow our robot team to operate under such uncertain conditions. As a result, meeting at bad areas lowers resilience to attacks in  terms of how long it takes the MRS to come to a resilient consensus. This is why we present the resilience results and the communication results in two different tables below. 
\end{remark}

\begin{table*}
\begin{center}
\renewcommand{\arraystretch}{1}
\captionof{table}{Resilient and Non-Resilient MIPP Statistics}\label{table:mipp_statistics}

\begin{tabular}{c c c c c c c c c c c c c c c}

\multicolumn{1}{c}{} & \multicolumn{2}{c}{$n=4$, $n_a=1$} & \multicolumn{1}{c}{} &  \multicolumn{2}{c}{$n=5$, $n_a=1$} & \multicolumn{1}{c}{} & \multicolumn{2}{c}{$n=6$, $n_a=1$} & \multicolumn{1}{c}{} & \multicolumn{2}{c}{$n=6$, $n_a=2$} & \multicolumn{1}{c}{} & \multicolumn{2}{c}{$n=7$, $n_a=2$}\\
\cline{2-3} \cline{5-6} \cline{8-9} \cline{11-12}  \cline{14-15}

 & $\mathcal{R}$ & $\mathcal{NR}$ & & $\mathcal{R}$ & $\mathcal{NR}$ & & $\mathcal{R}$ & $\mathcal{NR}$ & & $\mathcal{R}$ & $\mathcal{NR}$ & & $\mathcal{R}$ & $\mathcal{NR}$\\

$err(s_k)$   & $0.072$  & $0.866$ &     & $0.0135$ & $0.280$ &          & $0.012$ & $0.143$ &              & $0.0151$ & $0.9643$ &       & $0.013$ & $0.489$\\
$err(l_k)$   & $0.052$  & $3.114$ &     & $0.0320$ & $1.697$ &          & $0.026$ & $1.162$ &              & $0.0556$ & $4.0806$ &       & $0.013$ & $2.479$\\     
$err(y)$     & $2.864$  & $6.661$ &     & $2.8515$ & $5.534$ &          & $2.756$ & $4.433$ &              & $2.6064$ & $7.9577$ &       & $2.765$ & $5.775$\\

\end{tabular}

\end{center}
\end{table*}

\begin{table*}
\begin{center}
\renewcommand{\arraystretch}{1}
\captionof{table}{Communication Statistics}\label{table:coms_statistics}

\begin{tabular}{c c c c c c c c c c c c c c c}

\multicolumn{1}{c}{} & \multicolumn{2}{c}{$n=4$, $(2,2)$} & \multicolumn{1}{c}{} &  \multicolumn{2}{c}{$n=5$, $(2,2)$} & \multicolumn{1}{c}{} & \multicolumn{2}{c}{$n=6$, $(2,2)$} & \multicolumn{1}{c}{} & \multicolumn{2}{c}{$n=6$, $(3,3)$} & \multicolumn{1}{c}{} & \multicolumn{2}{c}{$n=7$, $(3,3)$}\\
\cline{2-3} \cline{5-6} \cline{8-9} \cline{11-12}  \cline{14-15}

 & $sb^*$ & $sb^r$ & & $sb^*$ & $sb^r$ && $sb^*$ & $sb^r$ && $sb^*$ & $sb^r$ && $sb^*$ & $sb^r$\\   


$P_r$    & $0.512$ & $0.214$ &       & $0.768$ & $0.344$ &      & $0.858$ & $0.415$ &          & $0.223$ & $0.059$ &     & $0.429$ & $0.185$ \\
$\#_{t}$ & $1.602$ & $3.842$ &       & $1.268$ & $2.586$ &      & $1.102$ & $2.531$ &          & $3.232$ & $5.259$ &     & $2.015$ & $4.23$ \\

\end{tabular}

\end{center}
\end{table*}

For each fixed number of robots and attackers, we generate 1000 random GPs that represent the unknown environment to be explored which is discretized into 625 possible sensing locations. Using Monte Carlo simulation, we run our resilient and non-resilient MIPP for each random GP. The environment is divided into $|\mathcal{AR}| = 3$ areas and each area is divided into $\mathcal{SB} = 10$ subareas similar to what was done in Figure \ref{fig:discrete_space}. The robots are provided random initial knowledge about the environment to induce different planning behaviors across the different simulations. Providing no initial knowledge results in robots planning the same paths initially, regardless of the unknown environment. Finally, attacked robots have their measurements corrupted by adding a random input $\epsilon_i$, sampled from a uniform distribution, to each measurement. The results are summarized in Tables \ref{table:mipp_statistics} and \ref{table:coms_statistics}.

\begin{figure*}[t!]%
    \centering
    \subfloat[Actual environmental GP]{{\includegraphics[width=4cm, draft = false]{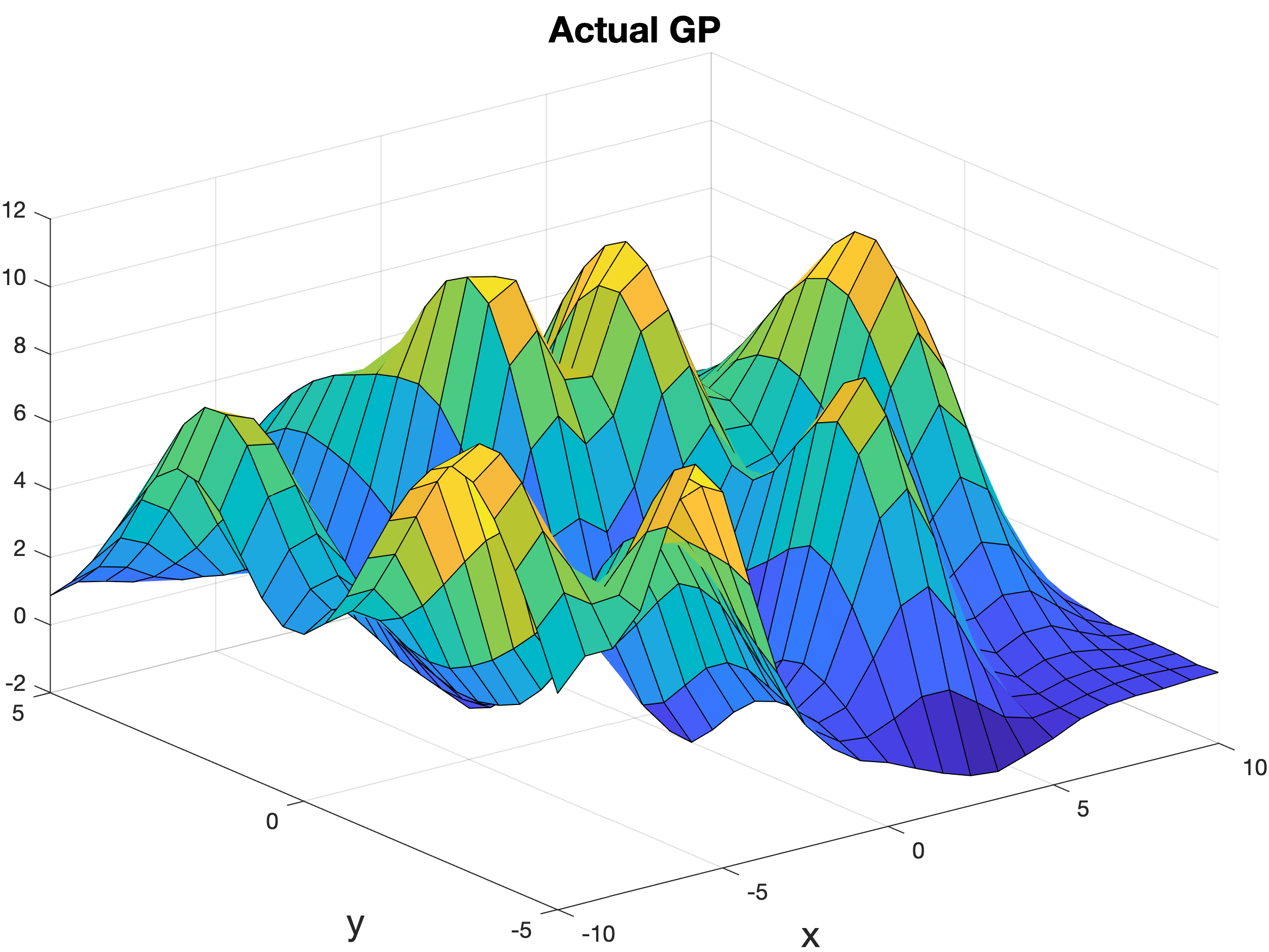} }\label{fig:actual_gp}}%
    \smallskip
    \hfill
    \subfloat[Initial knowledge of the GP]{{\includegraphics[width=4cm, draft = false]{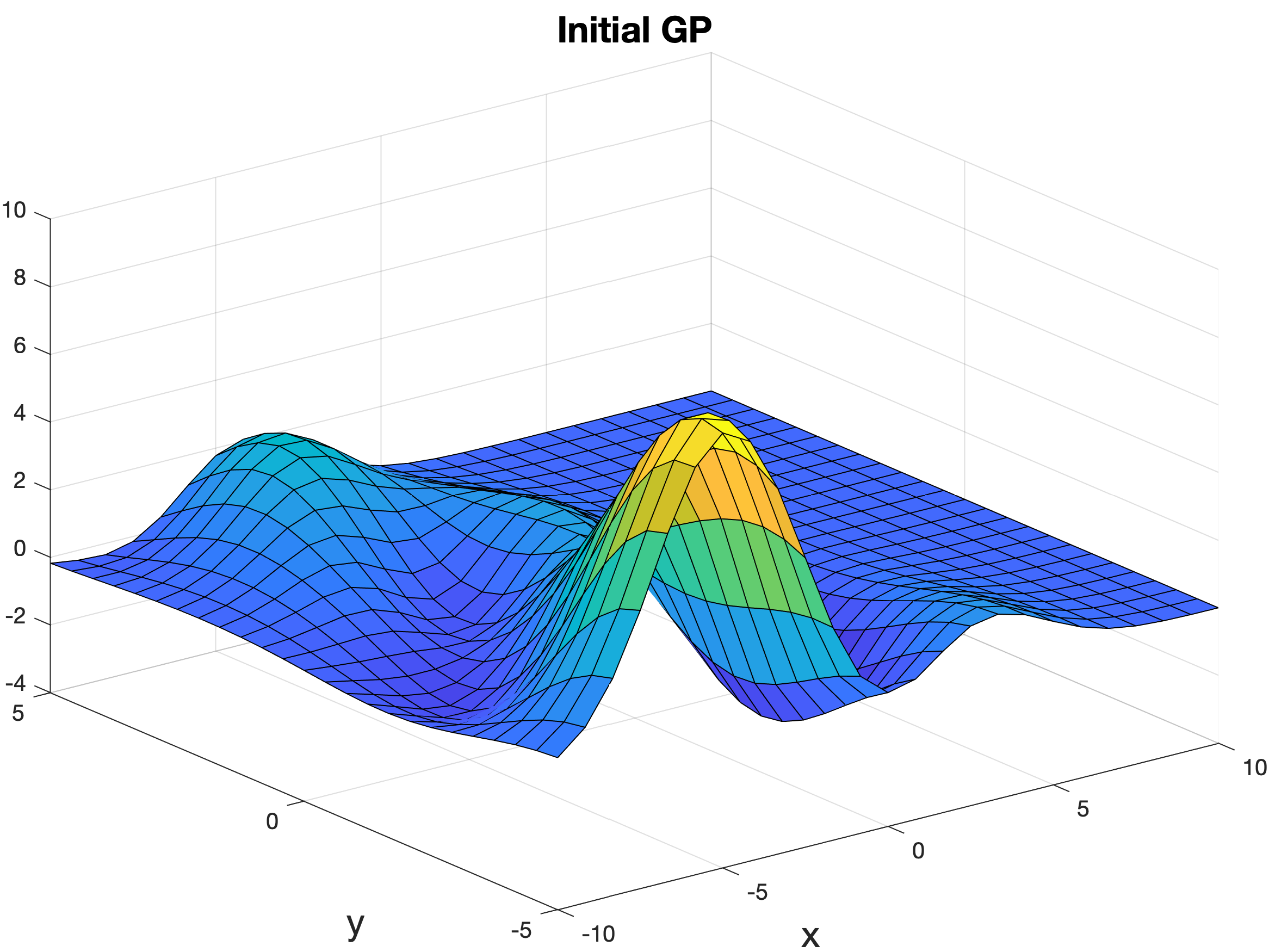} }\label{fig:initial_gp}}%
    \smallskip
    \hfill
    \subfloat[Learned GP by well-behaving robot using resilient MIPP]{{\includegraphics[width=4cm, draft = false]{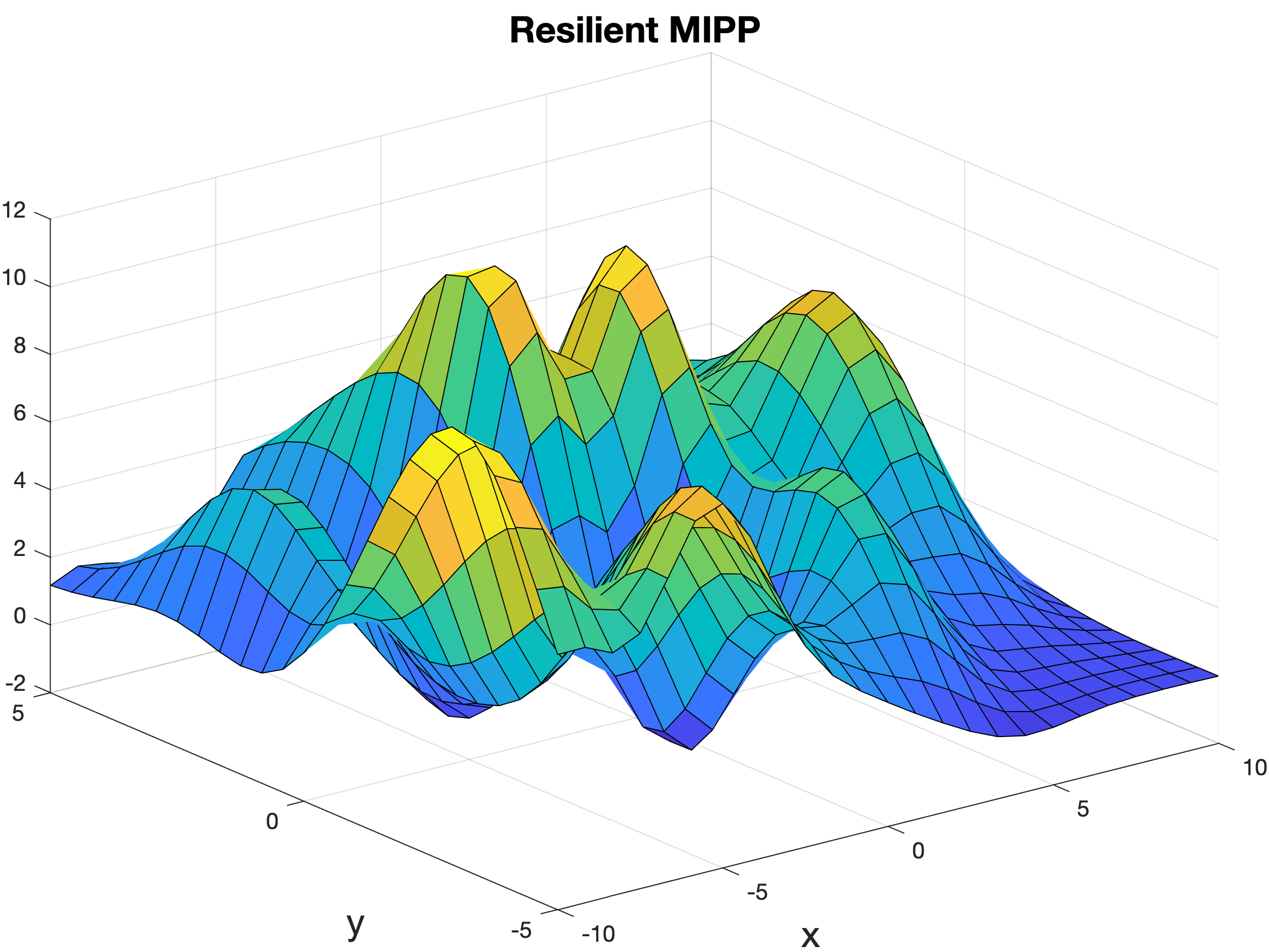} }\label{fig:res_gp}}%
    \smallskip
    \hfill
    \subfloat[Learned GP by well-behaving robot using non-resilient MIPP]{{\includegraphics[width=4cm, draft = false]{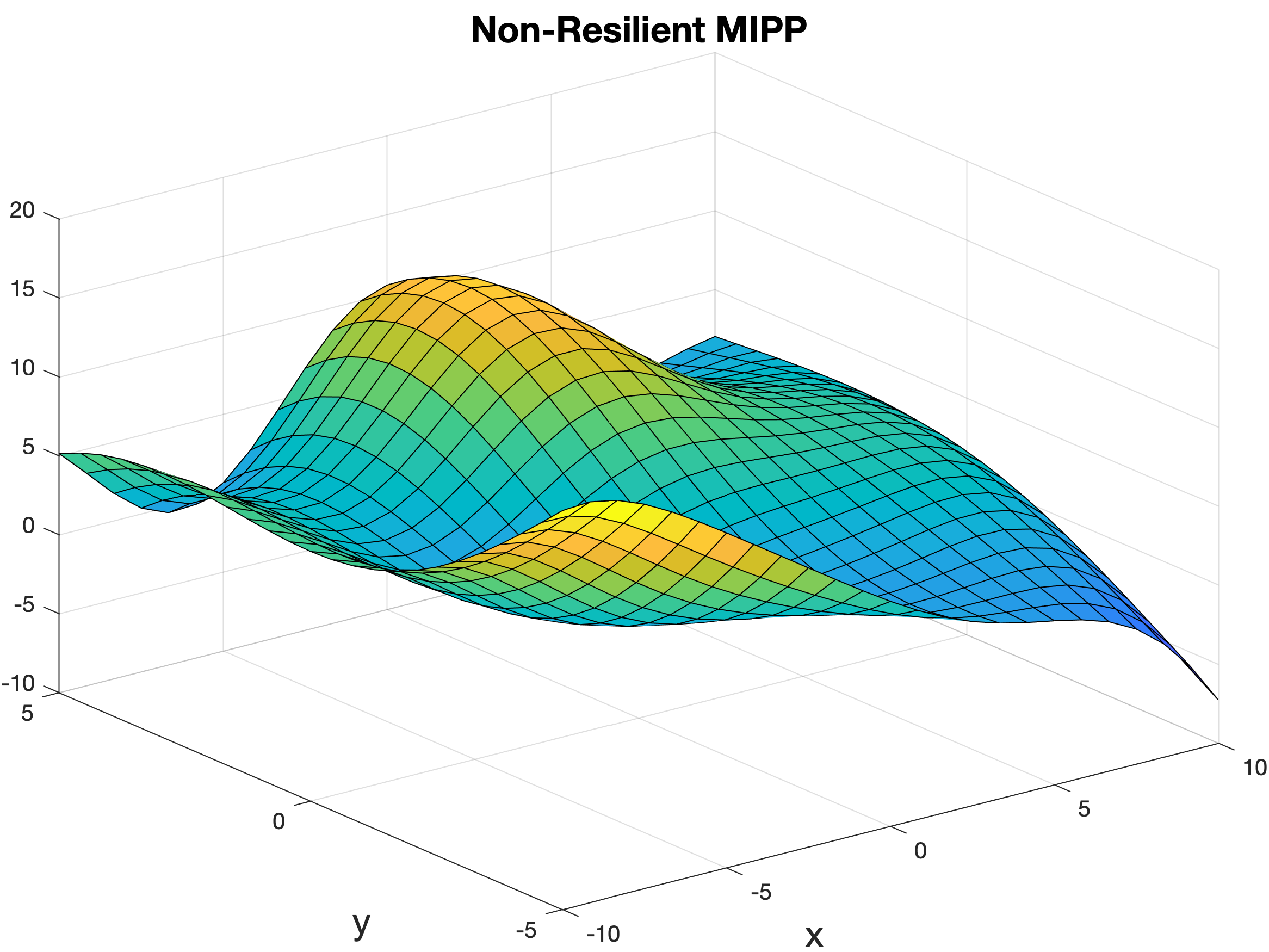} }\label{fig:nonres_gp}}%
    
        \caption{Comparing the learning ability of a well-behaving robot using resilient and non-resilient MIPP for $n=4$ and $n_a=1$}%
    \label{fig:compare_learn}%
\end{figure*}

Before elaborating on the results, let us first clarify the notations used in Table \ref{table:mipp_statistics}. As mentioned earlier, $n$ is the total number of robots and $n_a$ is the number of compromised robots. $\mathcal{R}$ refers to the results of the resilient MIPP algorithm which uses the W-MSR algorithm, while $\mathcal{NR}$ refers to the results of the non-resilient MIPP algorithm that uses the linear consensus expression given by \eqref{equ:linear_consen}. $err(s_k)$ refers to the absolute error between the actual and the learned parameter $s_k$. $err(l_k)$ refers to the absolute error between the actual and learned parameter $l_k$. Finally, $err(y)$ refers to the absolute error between the predicted and actual value of the Gaussian Process mean at every location of interest. All of these metrics represent the average over all runs and the average of well-behaving robots, meaning that these statistics represent what a single well-behaving robot learns on average. Looking at Table \ref{table:mipp_statistics}, we can see that regardless of the number of robots in the system and the number of attackers compromising the system, each well-behaving robot is able to learn the parameters of the covariance function with high accuracy and predict the environmental process with low error. 
On the other hand, we can clearly see that when running the non-resilient MIPP algorithm, well-behaving robots are significantly affected by the attackers in the system failing to both properly learn the correct hyper-parameters and make correct predictions. Figure \ref{fig:compare_learn} shows the result of a single run from the Monte Carlo simulation comparing the learning ability of a well-behaving robot using resilient and non-resilient MIPP.

Finally, we study the effects of probabilistic resilience on the number of communication rounds that have to occur between the robots and summarize the resuts in Table \ref{table:coms_statistics}. Note that $n, (r,s)$ refers to the number of robots and the amount of $(r,s)$-robustness imposed on the graph. Additionally, $sb^*$ denotes the subarea of highest probabilistic resilience, $sb^r$ is a random subarea chosen for comparison,  $P_r$ is the average probability of resilience over all runs, and $\#_t$ is the average number of communication rounds before the W-MSR algorithm can be applied. Clearly, selecting the subarea of highest probabilistic resilient is essential to minimize the number of communication rounds compared to meeting at a random location. Additionally, notice that for a fixed $(r,s)$, the probability of resilience of the MRS increases as the number of robots increases. As mentioned previously, we do not require the deterministic graph to be resilient in one instance because this will severely affect the number of communication rounds. Instead we only need the union of all the realized graphs over the communication rounds to be resilient \cite{prorock_reselience_2}. This is why for example, $P_r = 0.059$ for $n=6 ,(3,3)$ but only requires an average of $5.259$ rounds of communication instead of 17 rounds if we wait until a single realized instance is $(r,s)$-robust.

\section{Conclusion}

In this paper, we tackle the problem of resilient MIPP under the constraints of uncertainty in communication and the presence of corrupted robots. We model our environment using Gaussian Processes and use mutual information as a metric of informativeness. To model uncertainty, we associate every communication edge with a probability that represents the likelihood that the edge shows up in the graph. To robustify against corrupted robots, we use the W-MSR algorithm for resilient consensus. Our resilient MIPP approach consists of decoupling our multi-objective optimization into selecting meeting areas of high probabilistic resilient where robots unify their knowledge of the environment, and then selecting paths that maximize the mutual information during all other times. Finally, we show the validity of our results by comparing the learning ability of an MRS using a resilient and non-resilient MIPP algorithm.

\bibliography{arxiv.bib}
\bibliographystyle{IEEEtran}

\end{document}